\documentclass{article} 
\usepackage{iclr2027_conference,times}

\usepackage{hyperref}
\usepackage{url}
\usepackage{ulem}
\usepackage[utf8]{inputenc} 
\usepackage[T1]{fontenc}    
\usepackage{hyperref}       
\usepackage{url}            
\usepackage{booktabs}       
\usepackage{amsfonts}       
\usepackage{nicefrac}       
\usepackage{microtype}      
\usepackage{xcolor}         
\usepackage{amsmath,amsfonts,amsthm,amssymb}
\usepackage{mathtools}
\usepackage{placeins}
\usepackage{algorithm}
\usepackage[noend]{algpseudocode}

\usepackage{amsmath,amsfonts,bm}

\def\eqref#1{equation~\ref{#1}}

\def\1{\bm{1}}

\DeclareMathAlphabet{\mathsfit}{\encodingdefault}{\sfdefault}{m}{sl}
\SetMathAlphabet{\mathsfit}{bold}{\encodingdefault}{\sfdefault}{bx}{n}

\newcommand{\R}{\mathbb{R}}

\usepackage{graphicx}
\usepackage{hyperref}
\usepackage{url}
\usepackage{multirow}
\usepackage{subcaption}
\newtheorem{proposition}{Proposition}

\theoremstyle{definition}

\newtheorem*{asu*}{Assumption}

\usepackage{enumitem}

\renewcommand{\R}{\mathbb{R}}

\newcommand{\p}{{\rm I}\kern-0.18em{\rm P}}

\usepackage[utf8]{inputenc} 
\usepackage[T1]{fontenc}    
\usepackage{hyperref}       
\usepackage{url}            
\usepackage{booktabs}       
\usepackage{amsfonts}       
\usepackage{nicefrac}       
\usepackage{microtype}      
\usepackage{xcolor}         
\usepackage{parskip}

\title{Reconstruction-Optimized Expert Pruning for Mixture-of-Experts Language Models \\[0.01mm]}

\author{
\shortstack{
\textbf{Jelena Markovic-Voronov\textsuperscript{*,1}} \quad
\textbf{Petros Prastakos\textsuperscript{*,1,2}} \quad
\textbf{Kayhan Behdin\textsuperscript{*,1}} \quad
\textbf{Jincheng Cao\textsuperscript{1}}
\\[1.5mm]
\textbf{Zhipeng Wang\textsuperscript{3}} \quad
\textbf{Shayan Mohajer Hamidi\textsuperscript{1}} \quad
\textbf{Yuanda Xu\textsuperscript{1}}
\\[1.5mm]
\textbf{Zhengze Zhou\textsuperscript{1}} \quad
\textbf{Wenhui Zhu\textsuperscript{1}} \quad
\textbf{Rahul Mazumder\textsuperscript{1,2}}
\\[2mm]
{\normalfont
\textsuperscript{1}LinkedIn \qquad
\textsuperscript{2}MIT \qquad
\textsuperscript{3}Google}
}
}

\iclrfinalcopy 

\begin{document}

\maketitle

\begingroup
\renewcommand{\thefootnote}{}
\footnotetext{\textsuperscript{*}Equal contribution.}
\endgroup
\lhead{Preprint}

\begin{abstract}
Mixture-of-Experts (MoE) models reduce computation through sparse expert activation, but deployment still requires storing a large number of expert parameters. We study post-training expert pruning using task-domain calibration data as a subset-selection problem: given a pretrained MoE layer, select a fixed number of experts that best reconstruct its original output. Exact subset search is combinatorial and becomes intractable for modern MoE models with many experts per layer. We develop OptEP, a scalable continuous relaxation that directly optimizes differentiable expert-selection variables, followed by a lightweight reconstruction-based update of the retained expert weights. Across multiple MoE architectures and tasks spanning mathematical reasoning and code generation, our approach improves over the strongest existing expert-pruning baselines, with absolute accuracy gains of up to 15.05\% on Qwen3-30B-A3B, 14.55\% on Qwen3-235B-A22B, 11.06\% on GLM-4.5-Air, and 16.20\% on Mixtral-8x7B. The improvements are particularly pronounced under aggressive pruning, demonstrating that reconstruction-driven expert selection and weight adaptation can preserve substantially more of the original model's performance while reducing the number of retained experts.

\end{abstract}

\section{Introduction}
\label{sec:introduction}

Mixture-of-Experts (MoE) models increase model capacity without
proportionally increasing per-token computation by routing each token
to only a small subset of expert networks
\citep{shazeer2017outrageously,lepikhin2021gshard,fedus2022switch,du2022glam}.
This architecture underlies open-weight language models such as
Mixtral, DeepSeek-MoE, Qwen, and OLMoE
\citep{jiang2024mixtral,dai2024deepseekmoe,yang2025qwen3,muennighoff2025olmoe}.
However, the computational savings from sparse activation do not
eliminate the memory burden of storing the complete expert pool,
even though each token uses only a fraction of it. As modern MoEs
scale from tens of billions to trillions of parameters, this growing
parameter footprint poses a substantial deployment challenge,
motivating methods that reduce the number of experts while
preserving model capabilities.

Expert pruning addresses this challenge by permanently removing a
subset of experts after training. The central challenge is to identify
a smaller collection that preserves the capabilities of the original
model. Common strategies assign each expert an importance score
based on activation frequency, router probability, output magnitude,
or related local statistics, and retain the highest-scoring experts
\citep{muzio2024seer,dong2025domain,lasby2025reap}.
These approaches are simple and scalable, but selecting experts
individually according to their scores does not directly optimize
the quality of the retained subset as a whole.

The usefulness of an expert depends on the contributions of the
other experts retained alongside it. Experts with high individual
scores may provide overlapping contributions, while a lower-scoring
expert may supply complementary information that is difficult for
the others to reproduce. Consequently, a collection of individually
high-scoring experts need not be the collection that best preserves
the original layer's computation. This motivates a joint selection
criterion: evaluate how well the retained experts collectively
reconstruct the output of the full MoE layer on representative
calibration data.

Reconstruction-based selection makes this objective explicit, but
introduces a combinatorial optimization problem. Selecting $s$
experts from a pool of $E$ admits $\binom{E}{s}$ possible subsets.
Prior work evaluates candidate subsets exhaustively (over the $\binom{E}{s}$ pool) using layer-output
reconstruction error \citep{lu2024not}, an approach that is practical
for small expert pools but becomes prohibitive as the number of
experts per layer and/or the number of model layers grows. The challenge is therefore to make joint,
reconstruction-driven selection scalable without exhaustively
enumerating expert combinations.

Expert selection also leaves a second opportunity for improvement. After pruning,
the retained experts operate within a different expert pool from the one in which
they were pretrained, and their original parameters are not optimized to compensate
for the functions of removed experts. Adapting the retained experts after selection
can therefore recover performance that would otherwise be lost. This motivates
treating expert selection and post-pruning adaptation as two connected optimization
problems, both guided by reconstruction of the original layer output.

We propose \textbf{OptEP}, a two-stage framework that follows this principle. First, we formulate
expert pruning as layerwise subset selection and introduce a differentiable
reconstruction surrogate that jointly optimizes expert-selection variables.
This allows selection to account for the combined contribution of the retained
experts while avoiding exhaustive enumeration of expert subsets. Our flexible continuous optimization framework enables us to explore different forms of the reconstruction loss, that better capture the effect of infrequently activated experts through weighing different tokens in the calibration set. 
Second, after
fixing the retained subset of experts, we optimize its expert weights using a layerwise
reconstruction objective (with additional regularization).
Together, these stages address both which experts to retain and how to adapt
them after pruning, without requiring prohibitively expensive full-model retraining.

Our main contributions are:
\begin{itemize}[leftmargin=*, noitemsep]
    \item We formulate expert pruning as joint reconstruction-based subset selection
and develop a scalable differentiable optimization approach that directly optimizes layer-output
reconstruction without combinatorial enumeration. In addition, we study a token-weighted reconstruction loss that captures the effect of tokens routed through
infrequently activated experts.

    \item We introduce a reconstruction-based expert-weight update
    that adapts retained experts after pruning with the goal of recovering the accuracy lost from expert selection.

    \item We evaluate our framework across multiple MoE language
    models on mathematical reasoning and code-generation tasks.
    At the most aggressive pruning setting evaluated for each
    model--task pair, our method with expert-weight updates achieves absolute accuracy improvements of up to 15.05\% on Qwen3-30B-A3B, 14.55\% on Qwen3-235B-A22B, 11.06\% on GLM-4.5-Air, and 16.20\% on Mixtral-8x7B over the strongest competing baseline at the same retention budget. At moderate pruning levels, the pruned models retain most of their original performance, with substantial performance preservation even under aggressive pruning. For example, Qwen3-235B-A22B retains 97\% of its unpruned performance on a math benchmark after pruning 75\% of its experts and applying expert weight updates.

\end{itemize}

\section{Related Work}
\label{sec:related_work}

\paragraph{Importance-based expert pruning}
SEER-MoE \citep{muzio2024seer} prunes experts by activation frequency.
EASY-EP \citep{dong2025domain} scores experts on domain-specific
demonstrations by combining router weights and expert-output magnitudes
with token-importance
weights derived from input--output representation similarity.
Motivated by a reconstruction-error bound, REAP \citep{lasby2025reap}
averages router-weighted expert output magnitudes
over tokens activating each expert, reducing frequency bias against
experts with strong but infrequent contributions.
Shapley-based scoring \citep{huang2025discovering} accounts for
interactions through marginal contributions across sampled subsets.
These methods rank individual experts rather than directly optimizing
the reconstruction error of the retained set.

\paragraph{Reconstruction- and search-based expert selection}
NAEE \citep{lu2024not} exhaustively searches expert subsets to minimize
the discrepancy between original and pruned MoE-layer outputs on
calibration data. This is
tractable for Mixtral-8x7B's eight experts per layer, but scales
combinatorially to larger expert pools. EEP \citep{liu2024eep} uses
evolutionary search for pruning configurations and subsequent merging
weights. Our continuous relaxation optimizes a smooth reconstruction surrogate
with pretrained router and expert weights fixed, then discretizes the
selection to a fixed expert budget per layer.

\paragraph{Layer-adaptive pruning}
Prior work also considers non-uniform pruning across layers, using expert- and layer-level importance \citep{yang2024moei2,bai2025diep,yoon2026compel} or search over layer-wise pruning budgets \citep{liu2026evoesap}. In contrast, we apply uniform pruning, retaining the same number of experts in every layer.

\paragraph{Expert diversity and merging}
Clustering-based pruning preserves complementary experts by grouping
similar experts \citep{zhang2025diversifying} or comparing behavior
across task domains \citep{hu2026mosaic}.
Our merging baseline, M-SMoE \citep{li2023merge}, aligns expert neurons,
groups experts around frequently activated representatives using routing
similarity, and averages their parameters with activation-frequency weights. This consolidates knowledge into
fewer experts rather than simply discarding it. DERN \citep{zhou2025dern}
transfers knowledge through neuron-level recombination. We instead
retain an explicit expert subset and adapt its weights by reconstruction.

\paragraph{Post-pruning adaptation}
Prior recovery methods include regularization-based fine-tuning in SEER-MoE
\citep{muzio2024seer}, corrective fine-tuning during iterative pruning
\citep{jaiswal2025finding}, and Fisher-guided weight compensation in
COMPEL \citep{yoon2026compel}. Our separate recovery stage reconstructs
original layer outputs while regularizing retained experts toward
their pretrained weights. This connects to layerwise weight-pruning
methods such as Optimal Brain Compression, SparseGPT, and ALPS
\citep{obc,sparsegpt,alps}, while expert removal additionally changes
both the available functions and their routing weights. Broader compression
strategies also prune within experts or remove layers and blocks
\citep{he2024towards}.

\section{Method} \label{sec:method}

\paragraph{Background} Let $\mathcal{E}=\{1,\ldots,E\}$ denote the experts in
one MoE layer, $K\geq 1$ of which get activated for each token. Let
$\boldsymbol{\theta}_e$ denote the collection of weight matrices parameterizing
expert $e$, and let $f_e(\boldsymbol{x};\boldsymbol{\theta}_e)$ be the output
of expert $e$ for layer input activation $\boldsymbol{x}$ corresponding to a given token.  Let $\boldsymbol{\Theta}=\{\bm{\theta}_e\}_{e\in\mathcal{E}}$ denote the collection of all experts weights for the current layer.
We omit the layer index for notational simplicity; all quantities and optimization problems in this section are defined for a given layer.
For each token $\boldsymbol{x}$, the pretrained router produces routing scores
$\boldsymbol{a}(\boldsymbol{x})\in\R^E_{\geq 0}$.
Let $\mathcal{T}_K(\boldsymbol{a}(\boldsymbol{x}))\subseteq \mathcal{E}$ contain
the $K$ experts with the largest router scores for input $\boldsymbol{x}$ for a given token.
For all models considered in this work, the routing scores are normalized over the selected top-$K$ experts.  GLM-4.5-Air additionally contains one always-active shared expert.
We prune only the routed experts and keep the shared expert unchanged;
for notational simplicity, $\mathcal{E}$ refers only to the set of routed
experts throughout this section. Formally, the output of the original layer is given as

\begin{equation}
\label{eq:original_moe}
\boldsymbol{y}(\boldsymbol{x})
=
\sum_{e\in\mathcal{T}_K(\boldsymbol{a}(\boldsymbol{x}))}
p_e^{(K)}(\boldsymbol{x})f_e(\boldsymbol{x};\boldsymbol{\theta}_e),
\quad
p_e^{(K)}(\boldsymbol{x})
=
\frac{a_e(\boldsymbol{x})}
{\sum_{j\in\mathcal{T}_K(\boldsymbol{a}(\boldsymbol{x}))}a_j(\boldsymbol{x})}.
\end{equation}

We assume a calibration set contains $M$ sequences, each consisting of a prompt and completion. Let $T_m$ denote the length of calibration sequence $m$, for $1\leq m\leq M$, and let $\{\boldsymbol{x}_n^{(m)}\}_{n=1}^{T_m}$ denote the corresponding input activations to the layer for the $T_m$ calibration tokens used in the optimization.

\paragraph{Expert subset selection} We follow a layerwise pruning approach~\citep{obc}, pruning each MoE layer independently. Layerwise methods are widely used in LLM pruning~\citep{sparsegpt,wanda,alps}, offering high accuracy at substantially lower cost than end-to-end pruning. For a given layer, let $z_e\in\{0,1\}$ indicate whether expert $e\in\mathcal{E}$ is retained, where $z_e=1$ denotes retention and $z_e=0$ denotes expert removal. Let $\widehat{\boldsymbol{y}}^{(K)}(\boldsymbol{x};\boldsymbol{z},\boldsymbol{\Theta})$ denote the resulting layer output for token $\boldsymbol{x}$. Pruning according to $\boldsymbol{z}$ is equivalent to replacing the router scores $\boldsymbol{a}(\boldsymbol{x})$ with $\boldsymbol{z}\odot\boldsymbol{a}(\boldsymbol{x})$, where $\odot$ denotes element-wise multiplication. Thus, the pruned MoE layer output is given by

\begin{equation}
\label{eq:relaxed_pruned}
\widehat{ \boldsymbol{y}}^{(K)}(\boldsymbol{x};\boldsymbol{z},\boldsymbol{\Theta})=
\sum_{e\in\mathcal{T}_K(\boldsymbol{a}(\boldsymbol{x})\odot\boldsymbol{z})}
\pi_e(\boldsymbol{x};\boldsymbol{z})f_e(\boldsymbol{x};\boldsymbol{\theta}_e),
\quad \pi_e(\boldsymbol{x};\boldsymbol{z})
=
\frac{z_e a_e(\boldsymbol{x})}
{\sum_{j\in \mathcal{T}_K(\boldsymbol{a}(\boldsymbol{x})\odot\boldsymbol{{z}})} z_ja_j(\boldsymbol{x})}.
\end{equation}

Given a pruning budget $s$, with $K\leq s<E$, we retain exactly $s$ experts per layer by minimizing the reconstruction loss between the full and pruned models on the calibration set. We formulate the layerwise objective as the subset-selection problem: 

\begin{equation}
\label{eq:intro_subset_selection}
    \min_{\boldsymbol{z}}~~
    \frac{1}{2M}\sum_{m=1}^M
    \frac{1}{T_m}\sum_{n=1}^{T_m}
    \mathcal{L}\left(\boldsymbol{y}(\boldsymbol{x}_n^{(m)}),\widehat{ \boldsymbol{y}}^{(K)}(\boldsymbol{x}_n^{(m)};\boldsymbol{z},\boldsymbol{\Theta})
    \right) ~~ \text{s.t.}~~z_e\in\{0,1\}~\forall e\in\mathcal{E},~\sum_{e\in\mathcal{E}}z_e=s
\end{equation}
where $\mathcal{L}(\boldsymbol{u},\boldsymbol{v})$ quantifies the distance between two vectors. In particular, in this paper we assume $\mathcal{L}(\boldsymbol{u},\boldsymbol{v})=\|\boldsymbol{u}-\boldsymbol{v}\|_2^\alpha$, where $\alpha\in\{1,2\}$ is a hyperparameter. The \(1/T_m\) normalization gives each calibration sequence equal total weight, preventing longer sequences from dominating the reconstruction objective. Thus, \eqref{eq:intro_subset_selection} selects the $s$ experts that minimize aggregate layerwise reconstruction error over the calibration set, with $\boldsymbol{\Theta}$ fixed.

\paragraph{Continuous optimization of expert selection: OptEP~(uw)}
Directly optimizing~\eqref{eq:intro_subset_selection} requires a combinatorial search over expert subsets which is expensive for large expert pools.
To obtain a differentiable optimization formulation, we relax the binary selection variables to \(z\in[0,1]^E\) and replace the discontinuous Top-\(K\) pruned-layer output in \eqref{eq:relaxed_pruned} with a smooth surrogate
\begin{equation}
\label{eq:relaxed_router}
 \widetilde{\boldsymbol{y}}(\boldsymbol{x};\boldsymbol{z},\boldsymbol{\Theta})=
\sum_{e\in\mathcal{E}}
\tilde\pi_e(\boldsymbol{x};\boldsymbol{z})f_e(\boldsymbol{x};\boldsymbol{\theta}_e), ~~\text{where} ~~\tilde\pi_e(\boldsymbol{x};\boldsymbol{z})
=
\frac{z_e a_e(\boldsymbol{x})}
{\sum_{j\in\mathcal{E}} z_ja_j(\boldsymbol{x})}.
\end{equation}
This leads to the following continuous optimization problem:
\begin{equation}
\label{eq:continuous_pruning}
    \min_{\boldsymbol{z}}~~
    \frac{1}{2M}
    \sum_{m=1}^M
    \frac{1}{T_m}\sum_{n=1}^{T_m}
    \mathcal{L}\left(\boldsymbol{y}(\boldsymbol{x}_n^{(m)}),\widetilde{ \boldsymbol{y}}(\boldsymbol{x}_n^{(m)};\boldsymbol{z},\boldsymbol{\Theta})
    \right)~~~\text{s.t.}~~~
    z_e\in[0,1]~\forall e\in\mathcal{E}.
\end{equation}
We optimize \eqref{eq:continuous_pruning} while keeping
the pretrained router and expert weights $\bm\Theta$ fixed. We do not enforce the cardinality constraint $\sum_e z_e=s$ during continuous optimization; instead, after optimization, we discretize the solution by retaining the \(s\) experts with the largest values of \(z_e\). After discretization, the resulting model remains a standard \(K\)-sparsely activated MoE, with \(s\) rather than \(E\) experts per layer.

\paragraph{Weighted reconstruction objective: OptEP (w)}
In addition to the formulation in \eqref{eq:continuous_pruning}, we consider a weighted variant, OptEP (w), designed to reduce the dominance of frequently activated experts in the reconstruction objective. Tokens routed through infrequently used experts receive larger weights, giving these experts greater influence during selection and helping preserve specialized or complementary behavior that may otherwise be underrepresented in the calibration data.
Specifically, let $n_e$ denote the number of calibration tokens routed to expert $e$ by the unpruned model in the current layer. For a token with layer input $\boldsymbol{x}$, we define its weight as
\begin{equation} \label{eq:weights:sum}
    w(\bm x) = \sum\nolimits_{e\in\mathcal{T}_K(\boldsymbol{a}(\boldsymbol{x}))} {1}/{n_e},
\end{equation}
where we sum over the experts active for that token.
The weighted reconstruction objective based optimization problem is:
\begin{equation}
\label{eq:weighted_pruning}
    \min_{\boldsymbol{z}}~~
    \frac{1}{2M}
    \sum_{m=1}^M
    \frac{1}{T_m}\sum_{n=1}^{T_m}
    w(\bm x_n^{(m)}) \cdot \mathcal{L}\left(\boldsymbol{y}(\boldsymbol{x}_n^{(m)}),\widetilde{ \boldsymbol{y}}(\boldsymbol{x}_n^{(m)};\boldsymbol{z},\boldsymbol{\Theta})
    \right)~~\text{s.t.}~~z_e\in[0,1] \; \forall e\in\mathcal{E}.
\end{equation}

Appendix~\ref{sec:weighted_reconstruction_theory} provides a theoretical
interpretation of this weighting. Up to a constant normalization, it converts
the token-level reconstruction objective into an expert-balanced objective
that assigns equal relative weight to each expert-conditioned reconstruction
loss. This removes the direct dependence of the aggregation weights on expert
activation frequency and yields simple frequency-independent bounds under
reweightings of the expert-conditioned losses.

\paragraph{Reconstruction-based expert weight update: EWU}
To recover accuracy lost during pruning, we update the weights of the retained experts to compensate for those removed. Let $\boldsymbol{\Theta}'=\{\boldsymbol{\theta}'_e:
e\in\mathcal{E}, z_e=1\}$ denote the updated weights of the retained experts. With the selected subset $\boldsymbol{z}$ fixed, we optimize only the weights $\bm\theta'_e$ of selected experts ($z_e=1$) using the layerwise objective:
\begin{equation}
\label{eq:expert_weight_update}
    \min_{\boldsymbol{\Theta}'}~~
    \frac{1}{2M}
    \sum_{m=1}^M
    \frac{1}{T_m}\sum_{n=1}^{T_m}
    \mathcal{L}\left(\boldsymbol{y}(\boldsymbol{x}_n^{(m)}),\widehat{ \boldsymbol{y}}^{(K)}(\boldsymbol{x}_n^{(m)};\boldsymbol{z},\boldsymbol{\Theta}')
    \right)
+
\lambda
\sum_{\substack{e\in\mathcal{E}\\ z_e=1}}
\left\|\boldsymbol{\theta}'_e-\boldsymbol{\theta}_e\right\|_F^2,
\end{equation}
where $\lambda\geq 0$ is a pre-specified regularization parameter to avoid overfitting~\citep{sparsegpt,meng2024alps}. For a fixed layer, the router, retained expert subset, layer inputs are held fixed, and only the retained expert weights \(\boldsymbol{\Theta}'\) are optimized. The objective in~\eqref{eq:expert_weight_update} is differentiable in $\boldsymbol{\Theta}'$.

\paragraph{Layer Inputs for Expert Selection and EWU}
For the expert-selection objectives in \eqref{eq:continuous_pruning} and \eqref{eq:weighted_pruning}, $\bm{x}_n^{(m)}$ denotes the layer input obtained from a forward pass of the full, unpruned model. Expert selection is therefore performed independently across layers: pruning decisions made in one layer do not affect the inputs or expert selection in subsequent layers. We use this independent layerwise setup to allow comparisons with prior expert-pruning baselines, which likewise perform pruning independently across layers \citep{dong2025domain, lasby2025reap}.

In contrast, EWU is performed sequentially across layers. When optimizing a
given layer, $\bm{x}_n^{(m)}$ is obtained by forwarding the calibration data
through the pruned model, including the updated weights of all preceding
layers. Consequently, weight updates to earlier layers affect the inputs used
to optimize subsequent layers.

\paragraph{Joint versus fixed-score expert selection}
Importance-based pruning methods assign each expert a fixed score and retain
the highest-scoring experts. In Appendix~\ref{sec:joint_selection_theory}, we
show that such a fixed-score top-$s$ rule arises naturally from a local
first-order approximation of the OptEP reconstruction objective: the
resulting expert score is the negative gradient of the reconstruction loss
with respect to its selection variable at a fixed reference point. OptEP
instead optimizes the full nonlinear continuous reconstruction surrogate, so
these local scores evolve as the selection variables and reconstructed mixture
change. Hence the marginal value of an expert depends on the experts
currently emphasized, reducing the value of redundant experts while preserving
the value of complementary ones.


\section{Experiments}
\label{sec:experiments}

\paragraph{Experimental goals} We evaluate whether reconstruction-optimized expert selection improves downstream performance over existing expert-pruning methods across model scales, task domains, and pruning levels. We further study (i) whether inverse-frequency token weighting improves expert selection, (ii) whether reconstruction-based expert-weight updates recover additional performance after pruning, and (iii) whether our continuous relaxation can match the performance of exhaustive reconstruction-based subset selection when exact enumeration is feasible.

\subsection{Experimental Setup} \label{sec:experimental:setup}

\paragraph{Models}
We evaluate Qwen3-30B-A3B, Qwen3-235B-A22B~\citep{yang2025qwen3},
GLM-4.5-Air~\citep{zeng2025glm}, and Mixtral-8x7B-v0.1~\citep{jiang2024mixtral}.
The two Qwen models contain 128 routed experts in each of their 48 and 94
MoE layers, respectively, with Top-8 routing.
GLM-4.5-Air contains 46 transformer layers, of which the first is dense
and the remaining 45 are MoE layers, with 128 routed experts and one
always-active shared expert per MoE layer. It uses Top-8 routing; we prune
only the routed experts and leave the shared expert unchanged.
Mixtral-8x7B-v0.1 contains 8 experts in each of its 32 MoE layers with
Top-2 routing.

For each task domain (math or coding), we prune the same proportion of
routed experts from every MoE layer, producing a task-specific model with
the same number of layers as the original model but fewer routed experts
per layer. For Qwen, we retain 32, 48, and 64 experts per layer
(25\%, 37.5\%, and 50\%) for math, and 48, 64, and 96 experts
(37.5\%, 50\%, and 75\%) for coding. For GLM-4.5-Air, we retain
32, 48, and 64 routed experts per layer for math, and 64, 80, and 96
routed experts per layer for coding. For Mixtral-8x7B, we retain 4 or 6
of the 8 experts per layer (50\% or 75\%). We do not consider more
aggressive pruning settings because, for each model--task pair, retaining
fewer experts than the lowest setting reported above leads to substantial performance degradation.

\paragraph{Calibration}

We use domain-specific calibration for expert selection: models evaluated on math benchmarks are pruned using traces generated from math problems, while those evaluated on coding benchmarks use traces from code-generation problems~\citep{lucas2026reasoning}. The same base model is used to generate the calibration traces; this model is then pruned, and the resulting pruned model is used for final evaluation. For Qwen and GLM, we generate full-model calibration traces from OpenR1-Math-220k \citep{openr1} for math and Codeforces \citep{penedo2025codeforces} for coding. Each calibration sequence consists of the prompt concatenated with the model-generated completion, including its reasoning trajectory. 
Because Mixtral-8x7B performs poorly on MATH-500 and LiveCodeBench, these benchmarks provide limited resolution for comparing pruning methods. We therefore use GSM8K \citep{cobbe2021training} training examples for math and MBPP \citep{austin2021program} for coding, where differences between pruning methods are more readily measurable.

For all models and domains, we retain only untruncated samples for which the full model produces the correct answer. This ensures that expert selection is based on successful model trajectories rather than activations arising from failed reasoning or incorrect generations. Expert selection uses a fixed calibration-token budget; the sampled examples vary across seeds while the total number of calibration tokens remains fixed. The same calibration sequences used for expert selection are reused for the expert weight update, so both stages operate on an identical calibration distribution. Additional details on the calibration sets, including the token budget and number of sequences used, are provided in Appendix \ref{sec:calibration:details}.

\paragraph{Evaluation}

For Qwen and GLM, the pruned models selected using math calibration traces are evaluated zero-shot on MATH-500 \citep{hendrycks2021measuring, lightman2024let}, while those selected using coding calibration traces are evaluated on LiveCodeBench (LCB v4--v5) \citep{jain2025livecodebench}. We report mean pass@1 across seeds. Variation across seeds reflects both calibration-set sampling and stochastic evaluation sampling, with evaluation temperature fixed at 0.6. For Mixtral-8x7B, models pruned using GSM8K math calibration traces are evaluated on the GSM8K test set, while models pruned using MBPP coding calibration traces are evaluated on held-out MBPP examples. We report mean pass@1 across seeds. Because Mixtral-8x7B uses greedy decoding (temperature 0), there is no inference-time sampling variability. For GSM8K, variation across seeds therefore reflects only calibration-set selection; for MBPP, it additionally reflects variation in the seed-specific held-out evaluation split. Additional evaluation details, including evaluation set sizes and decoding settings, are provided in Appendix \ref{sec:evaluation:details}.

\paragraph{Baselines}

We compare against random pruning, routing-frequency pruning \citep{muzio2024seer},
EASY-EP \citep{dong2025domain}, REAP \citep{lasby2025reap} and NAEE \citep{lu2024not}.
These baselines, as well as our method, use \emph{uniform pruning}, retaining the same number of experts in every MoE layer. NAEE exhaustively enumerates candidate expert subsets, making it practical for Mixtral-8x7B, which has only 8 experts per layer, but computationally infeasible for the Qwen models with 128 experts per layer. We therefore evaluate NAEE only on Mixtral-8x7B.
We additionally compare against M-SMoE \citep{li2023merge}, which combines expert pruning with expert merging. 
We also include DiEP \citep{bai2025diep}, a non-uniform pruning method that allows the number of retained experts to vary across MoE layers. We were unable to successfully run the available DiEP implementation on the Qwen and GLM models due to implementation and scalability limitations, so we report DiEP results only for Mixtral-8x7B.
Additional details on the baselines are provided in Appendix~\ref{sec:baselines:details}.

For a given model, task, and retention level, all methods use the same calibration data and token budget and are evaluated under the same downstream protocol, controlling for differences in data, pruning budget, and evaluation setup.

\paragraph{Implementation and Compute Details}

We compare the base OptEP (uw) objective, its inverse-frequency-weighted variant OptEP (w), and their corresponding expert-weight-update variants against the baselines described above. The optimization details of these pruning and weight update methods are in Appendix \ref{sec:optimization:details}. Qwen3-235B-A22B and GLM-4.5-Air experiments used 8 H100 GPUs per seed, while Qwen3-30B-A3B and Mixtral-8x7B experiments used 4 H100 GPUs per seed.

\subsection{Main Results}

\paragraph{Results} Tables~\ref{tab:qwen30b-results}, \ref{tab:qwen235b-results}, \ref{tab:glm45air-results}, and \ref{tab:mixtral_results} compare the mean performance of OptEP (uw), OptEP (w), OptEP (w)~+ EWU, and the baseline methods. OptEP (w)~+ EWU applies the expert-weight update to the experts selected by OptEP (w). All reported values are averaged across experimental seeds. In each table, bold values indicate the best-performing method and underlined values indicate the second-best. Full results, including standard errors, are provided in Appendix~\ref{sec:additional:results}.

\paragraph{Pure Pruning Performance}

We treat OptEP (w) as our primary pruning method, as it performs better overall than the unweighted OptEP (uw) variant. Among methods that perform pure expert pruning without modifying expert weights, OptEP (w) achieves the best performance in 18 of 22 settings. Its advantages are largest under aggressive pruning, with improvements over the strongest prior pruning baseline of up to 10.45\% on Qwen3-30B-A3B, 10.32\% on Qwen3-235B-A22B, 4.60\% on GLM-4.5-Air, and 1.61\% on Mixtral-8x7B. All the reported differences are absolute percentage points. 
On Mixtral-8x7B, OptEP (w) performs best in all four settings, although the gains are smaller, potentially because the model has only 8 experts per layer and therefore offers less room for improved subset selection. In the four settings where OptEP (w) is outperformed by a prior baseline, the differences are at most 0.5\%.

More broadly, at least one of the two reconstruction-based variants OptEP (uw) or OptEP (w) outperforms the strongest prior pruning baseline in 21 of 22 settings. Overall, these results show that reconstruction-based expert selection performs strongly across models and tasks, with the largest gains under aggressive pruning.

At less aggressive pruning levels, the reconstruction-based methods closely approach the performance of the unpruned model. In some cases they even exceed it on average---e.g, on Qwen3-30B-A3B coding at \(s=96\), OptEP (w) achieves 61.44\% compared with 60.57\% for the full model.

\paragraph{Expert Weight Update Gains}

Applying expert weight updates after pruning improves performance in 19 of 22 settings compared with pruning alone.
On Qwen3-30B-A3B, OptEP (w) + EWU outperforms OptEP (w) in 5 of 6 settings, with gains of up to 4.60\% absolute improvement. On Qwen3-235B-A22B, OptEP (w)~+ EWU improves over OptEP (w) in all 6 settings, with gains of up to 4.23\%. On GLM-4.5-Air, OptEP (w) + EWU improves over OptEP (w) in 4 out of 6 settings, with gains up to 6.46\%.
On Mixtral-8x7B, OptEP (w)~+ EWU improves over OptEP (w) in all 4 settings, with gains of up to 15.27\%.

Across all four models, OptEP (w)~+ EWU achieves the best performance in 20 of 22 settings relative to the existing baselines. The absolute improvements reach up to 11.33\% on Qwen3-30B-A3B math, 15.05\% on Qwen3-30B-A3B coding, 10.60\% on Qwen3-235B-A22B math, 14.55\% on Qwen3-235B-A22B coding, 11.06\% on GLM-4.5-Air math, 6.34\% on GLM-4.5-Air coding,
16.20\% on Mixtral-8x7B math, and 10.80\% on Mixtral-8x7B coding.
The benefits of EWU are generally larger under more aggressive pruning, where removing experts induces a greater mismatch between the retained experts and the original layer output. This suggests that the reconstruction-based update helps the retained experts compensate for capacity lost during pruning, making expert selection and post-pruning adaptation complementary.

\begin{table}[H]
\centering
\small
\caption{Expert pruning results for Qwen3-30B-A3B on math (MATH-500 pass@1) and coding (LiveCodeBench pass@1). Results are mean accuracy in \% over three calibration seeds.}
\label{tab:qwen30b-results}
\begin{tabular}{lcccccc}
\toprule
\multirow{2}{*}{Method}
& \multicolumn{3}{c}{Math (MATH-500)}
& \multicolumn{3}{c}{Coding (LiveCodeBench)} \\
\cmidrule(lr){2-4}\cmidrule(lr){5-7}
& $s{=}32$ & $s{=}48$ & $s{=}64$
& $s{=}48$ & $s{=}64$ & $s{=}96$ \\
\midrule
Random
& $0.00$
& $0.00$
& $0.00$
& $0.00$
& $0.00$
& $0.00$ \\

Frequency
& $1.40$
& $19.27$
& $66.33$
& $0.00$
& $1.87$
& $\mathbf{61.44}$ \\

EASY-EP
& $31.73$
& \underline{$93.73$}
& $94.60$
& $17.54$
& $54.73$
& $61.07$ \\

REAP
& $76.07$
& $91.93$
& $94.73$
& $38.43$
& $52.49$
& $59.83$ \\

M-SMoE
& $1.33$
& $1.00$
& $1.07$
& $0.00$
& $0.00$
& $59.83$ \\
\midrule

OptEP (uw)
& $\mathbf{88.40}$
& $93.40$
& $94.67$
& \underline{$50.50$}
& \underline{$56.22$}
& $60.57$ \\

\textbf{OptEP (w)}
& $83.47$
& $93.53$
& \underline{$95.47$}
& $48.88$
& $54.60$
& $\mathbf{61.44}$ \\

\textbf{OptEP (w) + EWU}
& \underline{$87.40$}
& $\mathbf{94.40}$
& $\mathbf{96.13}$
& $\mathbf{53.48}$
& $\mathbf{56.59}$
& \underline{$61.19$} \\
\midrule\midrule

Full model (128 experts)
& \multicolumn{3}{c}{$96.40$}
& \multicolumn{3}{c}{$60.57$} \\

\bottomrule
\end{tabular}
\end{table}

\begin{table}[H]
\centering
\small
\caption{Expert pruning results for Qwen3-235B-A22B on math (MATH-500 pass@1) and coding (LiveCodeBench pass@1). Results are mean accuracy in \% over three calibration seeds.}
\label{tab:qwen235b-results}
\begin{tabular}{lcccccc}
\toprule
\multirow{2}{*}{Method}
& \multicolumn{3}{c}{Math (MATH-500)}
& \multicolumn{3}{c}{Coding (LiveCodeBench)} \\
\cmidrule(lr){2-4}\cmidrule(lr){5-7}
& $s{=}32$ & $s{=}48$ & $s{=}64$
& $s{=}48$ & $s{=}64$ & $s{=}96$ \\
\midrule
Random
& $0.67$
& $1.13$
& $0.60$
& $0.00$
& $0.00$
& $0.00$ \\

Frequency
& $0.07$
& $0.27$
& $93.20$
& $0.00$
& $45.52$
& $64.43$ \\

EASY-EP
& $71.33$
& $93.20$
& $95.20$
& $41.67$
& $56.97$
& $61.19$ \\

REAP
& $84.07$
& $93.60$
& $96.07$
& $41.17$
& $61.69$
& $64.68$ \\

M-SMoE
& $0.67$
& $0.13$
& $60.07$
& $0.00$
& $27.11$
& $64.68$ \\
\midrule

OptEP (uw)
& $87.80$
& $92.53$
& $95.53$
& $43.04$
& $62.31$
& \underline{$65.17$} \\

\textbf{OptEP (w)}
& \underline{$91.87$}
& \underline{$95.27$}
& \underline{$96.13$}
& \underline{$51.99$}
& \underline{$62.44$}
& $64.18$ \\

\textbf{OptEP (w) + EWU}
& $\mathbf{94.67}$
& $\mathbf{96.13}$
& $\mathbf{96.47}$
& $\mathbf{56.22}$
& $\mathbf{62.94}$
& $\mathbf{66.42}$ \\
\midrule\midrule

Full model (128 experts)
& \multicolumn{3}{c}{$97.60$}
& \multicolumn{3}{c}{$66.67$} \\

\bottomrule
\end{tabular}
\end{table}

\begin{table}[H]
\centering
\small
\caption{Expert pruning results for GLM-4.5-Air on math (MATH-500 pass\@1) and coding (LiveCodeBench pass\@1). Results are mean accuracy in \% over three calibration seeds.}
\label{tab:glm45air-results}
\begin{tabular}{lcccccc}
\toprule
\multirow{2}{*}{Method}
& \multicolumn{3}{c}{Math (MATH-500)}
& \multicolumn{3}{c}{Coding (LiveCodeBench)} \\
\cmidrule(lr){2-4}\cmidrule(lr){5-7}
& $s{=}32$ & $s{=}48$ & $s{=}64$
& $s{=}64$ & $s{=}80$ & $s{=}96$ \\
\midrule

Random
& $1.40$
& $5.13$
& $27.07$
& $0.50$
& $3.48$
& $22.39$ \\

Frequency
& $26.47$
& $56.93$
& $74.87$
& $3.23$
& $14.68$
& $32.21$ \\

EASY-EP
& $71.80$
& $92.07$
& $95.53$
& $45.77$
& $54.85$
& $\mathbf{64.30}$ \\

REAP
& $75.87$
& \underline{$93.73$}
& $96.60$
& $49.88$
& $59.33$
& \underline{$64.18$} \\

M-SMoE
& $2.47$
& $27.53$
& $67.27$
& $1.24$
& $9.33$
& $27.49$ \\
\midrule

OptEP (uw)
& $77.73$
& $92.53$
& $\mathbf{97.07}$
& $51.99$
& \underline{$62.56$}
& $62.06$ \\

\textbf{OptEP (w)}
& \underline{$80.47$}
& \underline{$93.73$}
& $96.20$
& \underline{$52.61$}
& $\mathbf{62.69}$
& $\mathbf{64.30}$ \\

\textbf{OptEP (w) + EWU}
& $\mathbf{86.93}$
& $\mathbf{94.93}$
& \underline{$97.00$}
& $\mathbf{56.22}$
& $61.82$
& $62.06$ \\
\midrule\midrule

Full model (128 experts)
& \multicolumn{3}{c}{$97.87$}
& \multicolumn{3}{c}{$65.30$} \\

\bottomrule
\end{tabular}
\end{table}

\begin{table}[H]
\centering
\small
\caption{Expert pruning results for Mixtral-8x7B-v0.1 on math (GSM8K flexible extraction) and coding (MBPP standard pass@1). Results are mean accuracy in \% over five seeds.}
\label{tab:mixtral_results}
\begin{tabular}{lcccc}
\toprule
\multirow{2}{*}{Method}
& \multicolumn{2}{c}{Math (GSM8K)}
& \multicolumn{2}{c}{Coding (MBPP)} \\
\cmidrule(lr){2-3}\cmidrule(lr){4-5}
& $s{=}4$ & $s{=}6$ & $s{=}4$ & $s{=}6$ \\
\midrule
Random
& $3.08$
& $20.15$
& $7.74$
& $40.25$ \\

Frequency
& $8.75$
& $8.96$
& $18.71$
& $37.51$ \\

EASY-EP
& $7.01$
& $25.26$
& $28.70$
& $47.57$ \\

REAP
& $11.45$
& $23.55$
& $25.07$
& $44.49$ \\

M-SMoE
& $10.63$
& \underline{$30.51$}
& $24.45$
& $38.05$ \\

NAEE brute-force
& $11.96$
& $24.96$
& $28.53$
& $43.87$ \\

DiEP
& 10.19
& 24.90
& 16.54
& 41.91 \\
\midrule

OptEP (uw)
& \underline{$14.04$}
& $24.58$
& $27.99$
& $47.48$ \\

\textbf{OptEP (w)}
& $12.89$
& $25.32$
& \underline{$30.31$}
& \underline{$47.61$} \\

\textbf{OptEP (w) + EWU}
& $\mathbf{28.16}$
& $\mathbf{32.90}$
& $\mathbf{39.50}$
& $\mathbf{50.19}$ \\
\midrule\midrule

Full model (8 experts)
& \multicolumn{2}{c}{$47.61$}
& \multicolumn{2}{c}{$59.34$} \\

\bottomrule
\end{tabular}
\end{table}

\subsection{Ablation Studies}

\paragraph{Effect of Token Weighting}
OptEP (w) improves over OptEP (uw) in 3 of 6 settings on Qwen3-30B-A3B, with gains of up to 0.87\%, and in 5 of 6 settings on Qwen3-235B-A22B, with substantially larger gains of up to 8.95\%. On GLM-4.5-Air, OptEP (w) improves over OptEP (uw) in 5 out of 6 settings, with gains up to 2.74\%.
On Mixtral-8x7B, OptEP (w) improves over OptEP (uw) in 3 of 4 settings, with gains of up to 2.32\%.
The largest drop from the unweighted to the weighted variant occurs on Qwen3-30B-A3B for both math and coding at the lowest sparsity levels; however, both variants still outperform existing baselines by substantial margins in these settings.

One possible explanation for the larger gains on Qwen3-235B-A22B is that, under the same fixed calibration-token budget, its longer generations result in fewer calibration sequences than for Qwen3-30B-A3B, as detailed in Appendix~\ref{sec:calibration:details}. The longer calibration sequences may cause frequently activated experts to dominate the unweighted reconstruction objective. The weighting partially counteracts this imbalance by assigning greater weight to tokens that activate less frequently used experts, potentially helping preserve experts that are important despite appearing less often in the calibration data. This suggests that the weighting may be particularly beneficial for models that produce long, verbose reasoning traces.

\paragraph{Alternative Token Weighting} In Appendix~\ref{sec:additional:results}, we report the full results for OptEP (w-alt), which uses a slightly different token-weighting scheme but otherwise follows the same approach as OptEP (w). Across the four models, the maximum absolute difference between OptEP (w) and OptEP (w-alt) is 3.20\% on GLM-4.5-Air, with differences below 2.6\% on the other four models. Overall, the two weighting schemes yield similar performance, suggesting that the benefits of inverse-frequency weighting are not highly sensitive to the precise weighting formulation.

\paragraph{Comparison with Brute-Force Subset Selection}
On Mixtral-8x7B, OptEP (w) achieves accuracy comparable to or higher than NAEE, with gains of up to 3.74\%. NAEE \citep{lu2024not} performs exhaustive search over expert subsets using a reconstruction objective similar to ours, but with two differences: it does not normalize the loss per sequence, and it computes the Frobenius-norm reconstruction error over batches of calibration sequences, whereas our objective computes the $\ell_2$ reconstruction error over individual calibration tokens.

To provide a more direct comparison, Table \ref{tab:mixtral_results_full} in Appendix~\ref{sec:additional:results} additionally reports two brute-force variants that optimize the exact discrete counterparts of our objectives. OptEP (uw) brute force exactly solves the subset-selection problem in \eqref{eq:intro_subset_selection}, while OptEP (w) brute force exactly solves the corresponding weighted objective. Across the evaluated settings, OptEP (uw) and OptEP (w) perform similarly to their respective brute-force counterparts. Across the evaluated Mixtral settings, OptEP (uw) differs from its exact brute-force counterpart by at most 2.96\%, while OptEP (w) differs from its corresponding brute-force solution by at most 2.53\% absolute difference. These results suggest that the proposed continuous surrogate provides a useful approximation to the underlying discrete subset-selection objectives while avoiding the combinatorial cost of exhaustive search.

\paragraph{Expert Weight Updates Across Pruning Variants}

To assess whether EWU depends on the expert-selection objective, we apply it after both OptEP (uw) and OptEP (w) across all evaluated models in Appendix~\ref{sec:additional:results}. EWU improves downstream performance in 20 of 22 settings when applied after OptEP (uw) and in 19 of 22 settings when applied after OptEP (w), indicating that its benefits are not specific to the weighted selection objective. In most cases, the pruning method with stronger pre-update performance also performs better after EWU, although this relationship does not hold universally. When the pre-update performances are close, EWU can change their relative ordering, whereas larger gaps after pruning are generally preserved after the update. More broadly, EWU can be applied after any expert-selection baseline, although substantial performance gaps introduced during pruning are unlikely to be reversed by the subsequent weight update.

\section{Limitations}
\label{sec:limitations}

Our experiments focus on four open-weight MoE model families and on mathematical reasoning and code-generation tasks, so the extent to which the results generalize to other architectures and domains remains to be evaluated. We additionally use domain-specific calibration data consisting of correct trajectories generated by the full model; performance under noisier, unlabeled, or out-of-domain calibration data remains to be studied. We use uniform pruning, retaining the same number of experts in every layer. Allowing layer-specific pruning budgets could provide further improvements though selecting these layer-specific pruning budgets remain an interesting research direction. Although our continuous optimization based formulation avoids the combinatorial subset enumeration required by exact reconstruction-based pruning, expert selection and the subsequent weight-update stage still require per-layer optimization, introducing additional one-time pruning cost compared with simpler importance-based heuristics which in our experiments generally yield lower accuracy.

\section{Conclusion}
\label{sec:conclusion}

We presented a reconstruction-optimized framework for post-training expert pruning in Mixture-of-Experts language models. Rather than ranking experts independently, our approach formulates pruning as joint subset selection and uses a continuous reformulation to make reconstruction-based expert selection scalable to MoE models with large expert pools. We further introduced an inverse-frequency-weighted reconstruction objective and a lightweight reconstruction-based expert-weight update that adapts the retained experts after pruning. Across Qwen3-30B-A3B, Qwen3-235B-A22B, GLM-4.5-Air, and Mixtral-8x7B on mathematical reasoning and code-generation benchmarks, reconstruction-based expert selection improves over existing pruning methods in most evaluated settings, particularly under aggressive pruning, while the subsequent weight update recovers additional performance. When exhaustive subset selection is feasible, our continuous optimization based approach also achieves comparable performance. Overall, these results support that jointly optimizing the retained expert subset based on reconstruction quality, rather than selecting experts independently using local importance scores, is a promising, scalable approach to MoE compression.

\section{Acknowledgement}
Petros Prastakos contributed to this work while he was an intern at LinkedIn during summer 2026. This
work is not a part of his MIT research. Rahul Mazumder contributed to this work while he was a
consultant for LinkedIn (in compliance with MIT’s outside professional activities policies). This
work is not a part of his MIT research.
Zhipeng Wang contributed to the work while at LinkedIn.

\bibliography{ref}
\bibliographystyle{iclr2027_conference}

\appendix

\section{Theoretical Motivation for Weighted Reconstruction}
\label{sec:weighted_reconstruction_theory}

We provide an expert-balanced interpretation of the proposed weighted
reconstruction objective.

Consider an MoE layer with $E$ experts and a calibration set containing
$N=\sum_{m=1}^{M}T_m$ tokens. Let $\mathcal T(t)$ denote the set of $K$
experts activated by token $t$ in the original model, where
$|\mathcal T(t)|=K$. For expert $e$, define
\begin{equation}
    \mathcal X_e
    =
    \{t:e\in\mathcal T(t)\},
    \qquad
    n_e
    =
    |\mathcal X_e|.
    \label{eq:expert_token_set}
\end{equation}
Thus, $n_e$ denotes the activation frequency of expert $e$ on the calibration
set. We assume $n_e>0$ for every expert considered below, so that the
inverse-frequency weights are well-defined.

To account for the per-sequence length normalization used in
Eqs.~\eqref{eq:continuous_pruning} and~\eqref{eq:weighted_pruning},
let
\begin{equation}
    \ell_t(\mathbf z)
    :=
    \frac{N}{2M T_{m(t)}}
    \mathcal L
    \bigl(
        y_t,\widetilde y_t(\mathbf z)
    \bigr),
    \label{eq:normalized_token_loss}
\end{equation}
where $m(t)$ denotes the calibration sequence containing token $t$.
With this definition, the unweighted reconstruction objective can be written
exactly as
\begin{equation}
    L_{\mathrm{unweighted}}
    =
    \frac{1}{N}
    \sum_{t=1}^{N}
    \ell_t.
    \label{eq:unweighted_flattened}
\end{equation}

For simplicity, we suppress the dependence of $\ell_t$ on $\mathbf z$ below.
Define the \emph{expert-conditioned reconstruction loss} as
\begin{equation}
    L_e
    =
    \frac{1}{n_e}
    \sum_{t\in\mathcal X_e}
    \ell_t.
    \label{eq:expert_conditioned_loss}
\end{equation}
Importantly, $L_e$ is not the reconstruction error attributable solely to
expert $e$. Rather, it is the average sequence-normalized reconstruction loss
over calibration tokens for which expert $e$ is activated.

\subsection{From Token Averaging to Expert-Balanced Reconstruction}

The following proposition characterizes how inverse-frequency token weighting
changes the aggregation of expert-conditioned reconstruction losses.

\begin{proposition}[Expert-balanced interpretation of inverse-frequency weighting]
\label{prop:expert_balanced_weighting}
Let
$
    p_e
    :=
    \frac{n_e}{KN}.
$
Then $\sum_{e=1}^{E}p_e=1$, and the unweighted reconstruction objective can
be written as
\begin{equation}
    L_{\mathrm{unweighted}}
    =
    \sum_{e=1}^{E}
    p_e L_e.
    \label{eq:unweighted_expert_average}
\end{equation}
Thus, the unweighted objective aggregates expert-conditioned losses in
proportion to their calibration-time activation frequencies.

In contrast, for the proposed token weight
$
    w_t
    =
    \sum_{e\in\mathcal T(t)}
    \frac{1}{n_e},
    \label{eq:appendix_inverse_frequency_weight}
$
the weighted reconstruction objective satisfies
\begin{equation}
    L_{\mathrm{weighted}}
    :=
    \frac{1}{N}
    \sum_{t=1}^{N}
    w_t\ell_t
    =
    \frac{1}{N}
    \sum_{e=1}^{E}
    L_e.
    \label{eq:weighted_expert_sum}
\end{equation}
Consequently, up to the constant factor $E/N$, minimizing the weighted
objective is equivalent to minimizing the uniform expert average
$
    \frac{1}{E}
    \sum_{e=1}^{E}
    L_e.
    \label{eq:expert_balanced_objective}
$
\end{proposition}

\begin{proof}
Since every token activates exactly $K$ experts,
\begin{equation}
    \sum_{e=1}^{E}
    \sum_{t\in\mathcal X_e}
    \ell_t
    =
    K
    \sum_{t=1}^{N}
    \ell_t.
\end{equation}
Therefore,
\begin{align}
    L_{\mathrm{unweighted}}
    &=
    \frac{1}{N}
    \sum_{t=1}^{N}\ell_t =
    \frac{1}{KN}
    \sum_{e=1}^{E}
    \sum_{t\in\mathcal X_e}
    \ell_t =
    \sum_{e=1}^{E}
    \frac{n_e}{KN}L_e
    =
    \sum_{e=1}^{E}
    p_eL_e.
\end{align}

For the weighted objective,
\begin{align}
    \sum_{t=1}^{N}w_t\ell_t
    &=
    \sum_{t=1}^{N}
    \sum_{e\in\mathcal T(t)}
    \frac{1}{n_e}\ell_t =
    \sum_{e=1}^{E}
    \frac{1}{n_e}
    \sum_{t\in\mathcal X_e}
    \ell_t =
    \sum_{e=1}^{E}L_e.
\end{align}
Dividing by $N$ gives
Eq.~\eqref{eq:weighted_expert_sum}. Since $E/N$ is independent of the selection variables, minimizing this
objective is equivalent to minimizing the uniform expert average defined
above.
\end{proof}

Proposition~\ref{prop:expert_balanced_weighting} gives a direct interpretation
of the proposed weighting. Although the standard reconstruction objective
assigns the prescribed sequence-normalized weight to each token, when viewed
through expert-conditioned losses it weights expert $e$ proportionally to its
activation frequency $p_e$. The proposed inverse-frequency weighting removes
this frequency dependence and gives every expert-conditioned loss equal
relative weight.

\subsection{A Frequency-Independent Bound on Expert-Conditioned Loss}

The expert-balanced formulation also yields a simple consequence. Since
$L_e\geq0$ for every expert,
\begin{equation}
    L_e
    \leq
    \sum_{j=1}^{E}L_j
    =
    E
    \left(
        \frac{1}{E}\sum_{j=1}^{E}L_j
    \right),
    \qquad
    \forall e.
\end{equation}
Hence,
\begin{equation}
    \max_e L_e
    \leq
    E
    L_{\mathrm{bal}},
    \qquad
    L_{\mathrm{bal}}
    :=
    \frac{1}{E}\sum_{e=1}^{E}L_e.
    \label{eq:weighted_worst_expert}
\end{equation}

By comparison, Eq.~\eqref{eq:unweighted_expert_average} gives
\begin{equation}
    p_eL_e
    \leq
    L_{\mathrm{unweighted}},
\end{equation}
and therefore
\begin{equation}
    L_e
    \leq
    \frac{1}{p_e}
    L_{\mathrm{unweighted}}
    =
    \frac{KN}{n_e}
    L_{\mathrm{unweighted}}.
\end{equation}

Thus, at the level of these elementary bounds, the bound associated with the
expert-balanced objective does not explicitly deteriorate with expert
activation frequency, whereas the corresponding bound obtained from the
unweighted objective scales as $1/p_e$. 

\subsection{Robustness to Reweighting of Expert-Conditioned Losses}

A related interpretation follows by considering different relative weights on
the expert-conditioned losses. Let
\[
    \mathbf q=(q_1,\ldots,q_E),
    \qquad
    q_e\geq0,
    \qquad
    \sum_{e=1}^{E}q_e=1,
\]
and define
\begin{equation}
    L_{\mathbf q}
    =
    \sum_{e=1}^{E}q_eL_e.
\end{equation}

For the unweighted objective,
\begin{align}
    L_{\mathbf q}
    &=
    \sum_{e=1}^{E}
    \frac{q_e}{p_e}
    p_eL_e \leq
    \left(
        \max_e\frac{q_e}{p_e}
    \right)
    L_{\mathrm{unweighted}}.
    \label{eq:unweighted_reweighted_bound}
\end{align}
This bound can become loose when an expert has small calibration frequency
$p_e$ but receives substantially larger relative weight under $\mathbf q$.

For the expert-balanced objective,
\begin{equation}
    L_{\mathbf q}
    \leq
    \max_eL_e
    \leq
    E L_{\mathrm{bal}}.
    \label{eq:balanced_reweighted_bound}
\end{equation}
Equivalently,
\begin{equation}
    \sup_{\mathbf q\in\Delta_E}
    L_{\mathbf q}
    =
    \max_eL_e
    \leq
    E L_{\mathrm{bal}},
\end{equation}
where $\Delta_E$ denotes the probability simplex over experts.

Thus, for fixed expert-conditioned losses $\{L_e\}$, the expert-balanced
formulation provides a bound that does not depend on the calibration
frequencies $\{p_e\}$. This gives a robustness interpretation with
respect to changing the relative weights assigned to expert-conditioned
losses. We emphasize that this argument considers reweighting of fixed
$\{L_e\}$ and does not by itself characterize general distribution shift,
under which the expert-conditioned losses may also change.

\paragraph{Connection to frequency-normalized expert pruning}
The above motivation is related to the frequency-normalization principle used in prior MoE pruning methods such as REAP. Such methods seek to avoid conflating an expert's functional importance with how frequently it is activated. Our weighting applies a similar principle directly to the reconstruction objective: instead of allowing frequently activated experts to dominate token-level reconstruction, we normalize their contribution such that each expert-conditioned reconstruction loss receives equal weight.

\section{Joint Reconstruction Optimization versus Fixed-Score Expert Selection}
\label{sec:joint_selection_theory}

In this section, we connect our reconstruction-based optimization to
importance-score expert pruning. We show that a local first-order
approximation of the OptEP objective reduces exactly to a fixed-score
top-$s$ expert-selection rule. The resulting score is the local negative
gradient of the reconstruction objective with respect to each expert-selection
variable. The full OptEP optimization can therefore be viewed as
generalizing a fixed-score rule by continually updating these local scores as
the reconstructed mixture changes.

\subsection{Local Reconstruction Scores}

For notational simplicity, we flatten the calibration sequences into tokens
indexed by $t$. Let $x_t$ denote the layer input activation of token $t$, and
define
\[
    a_{t,e}=a_e(x_t),
    \qquad
    f_{t,e}=f_e(x_t;\theta_e),
    \qquad
    y_t=y(x_t).
\]
Since the pretrained router and expert weights $\Theta$ are fixed during
expert selection, we write
\begin{equation}
    \widetilde y_t(\mathbf z)
    :=
    \widetilde y(x_t;\mathbf z,\Theta)
    =
    \frac{
        \sum_{e=1}^{E} z_e a_{t,e} f_{t,e}
    }{
        \sum_{e=1}^{E} z_e a_{t,e}
    }.
    \label{eq:appendix_continuous_output}
\end{equation}

Define
\begin{equation}
    D_t(\mathbf z)
    =
    \sum_{j=1}^{E} z_j a_{t,j},
\end{equation}
and assume $D_t(\mathbf z)>0$ for the selection vectors considered below.

Let $c_t$ denote the coefficient of token $t$ in the continuous
reconstruction objective in Eq.~\eqref{eq:continuous_pruning}; specifically,
$c_t=1/(2MT_{m(t)})$ for a token belonging to calibration sequence $m(t)$.
The reconstruction objective can therefore be written as
\begin{equation}
    R(\mathbf z)
    =
    \sum_t
    c_t
    \mathcal L
    \bigl(
        y_t,\widetilde y_t(\mathbf z)
    \bigr).
    \label{eq:appendix_reconstruction_objective}
\end{equation}

Differentiating the normalized surrogate output with respect to $z_e$ gives
\begin{align}
    \frac{\partial \widetilde y_t(\mathbf z)}
         {\partial z_e}
    &=
    \frac{
        a_{t,e} f_{t,e} D_t(\mathbf z)
        -
        a_{t,e}
        \sum_j z_j a_{t,j}f_{t,j}
    }{
        D_t(\mathbf z)^2
    } =
    \frac{a_{t,e}}{D_t(\mathbf z)}
    \left(
        f_{t,e}
        -
        \widetilde y_t(\mathbf z)
    \right).
    \label{eq:output_selection_gradient}
\end{align}

Consequently, wherever $\mathcal L$ is differentiable,
\begin{equation}
    \frac{\partial R(\mathbf z)}{\partial z_e}
    =
    \sum_t
    c_t
    \frac{a_{t,e}}{D_t(\mathbf z)}
    \left\langle
        \nabla_2
        \mathcal L
        \bigl(
            y_t,\widetilde y_t(\mathbf z)
        \bigr),
        f_{t,e}-\widetilde y_t(\mathbf z)
    \right\rangle.
    \label{eq:general_selection_gradient}
\end{equation}

Equation~\eqref{eq:general_selection_gradient} gives a local importance score
for each expert under the continuous reconstruction objective. In particular,
the local marginal effect of expert $e$ depends not only on its own router
score and output, but also on the current reconstructed mixture through
$D_t(\mathbf z)$ and $\widetilde y_t(\mathbf z)$.

For the squared reconstruction loss
\[
    \mathcal L(y,\widetilde y)
    =
    \|y-\widetilde y\|_2^2,
\]
define the local marginal reconstruction score
\begin{equation}
    S_e(\mathbf z)
    :=
    -\frac{\partial R(\mathbf z)}{\partial z_e}.
\end{equation}
Using Eq.~\eqref{eq:general_selection_gradient}, we obtain
\begin{equation}
    S_e(\mathbf z)
    =
    \frac{1}{M}
    \sum_t
    \frac{1}{T_{m(t)}}
    \frac{a_{t,e}}{D_t(\mathbf z)}
    \left\langle
        y_t-\widetilde y_t(\mathbf z),
        f_{t,e}-\widetilde y_t(\mathbf z)
    \right\rangle.
    \label{eq:marginal_reconstruction_score}
\end{equation}

The inner product in Eq.~\eqref{eq:marginal_reconstruction_score} compares
the current reconstruction residual,
$y_t-\widetilde y_t(\mathbf z)$, with the direction in which increasing
$z_e$ moves the reconstructed mixture,
$f_{t,e}-\widetilde y_t(\mathbf z)$. Thus, an expert has a large positive
local score when increasing its selection weight moves the current mixture
in a direction that reduces the reconstruction residual.

\subsection{First-Order Approximation Yields Fixed-Score Selection}

The connection to fixed-score pruning can be made explicit by locally
linearizing the reconstruction objective. Consider a reference point
$\mathbf z^{(0)}$ at which $R$ is differentiable and
$D_t(\mathbf z^{(0)})>0$ for every calibration token. A first-order Taylor
expansion gives
\begin{equation}
    R(\mathbf z)
    =
    R(\mathbf z^{(0)})
    +
    \nabla R(\mathbf z^{(0)})^\top
    (\mathbf z-\mathbf z^{(0)})
    +
    o\bigl(
        \|\mathbf z-\mathbf z^{(0)}\|
    \bigr).
\end{equation}

Define the local expert score
\begin{equation}
    s_e^{(0)}
    :=
    -
    \left.
    \frac{\partial R(\mathbf z)}
         {\partial z_e}
    \right|_{\mathbf z=\mathbf z^{(0)}}.
\end{equation}
Ignoring terms independent of $\mathbf z$ and higher-order terms, the local
linear model becomes
\begin{equation}
    R_{\mathrm{lin}}(\mathbf z)
    =
    -
    \sum_{e=1}^{E}
    s_e^{(0)} z_e
    +
    \mathrm{const}.
    \label{eq:local_linear_objective}
\end{equation}

\begin{proposition}[First-order approximation yields fixed-score selection]
\label{prop:first_order_score}
Let $\mathbf z^{(0)}$ be a reference point satisfying the conditions above.
Minimizing the first-order approximation
in Eq.~\eqref{eq:local_linear_objective} subject to
\[
    \mathbf z\in\{0,1\}^{E},
    \qquad
    \sum_{e=1}^{E}z_e=s,
\]
is exactly equivalent to retaining the $s$ experts with the largest local
scores $s_e^{(0)}$.
\end{proposition}

\paragraph{Interpretation as a frozen-gradient pruning rule}
Proposition~\ref{prop:first_order_score} shows that fixed-score expert
selection arises naturally as a local linearization of our reconstruction
objective. At a reference mixture $\mathbf z^{(0)}$, the score
$s_e^{(0)}$ measures the instantaneous decrease in reconstruction error
associated with increasing the selection weight of expert $e$. If these
scores are computed once and then frozen, expert selection reduces exactly
to a standard top-$s$ importance-ranking rule.

This provides a structural connection between reconstruction optimization
and score-based pruning. The first-order model replaces the nonlinear
reconstruction objective by a single additive expert score computed at one
reference mixture.

\subsection{Dynamic Scores under Joint Reconstruction Optimization}

The full OptEP optimization differs from the first-order approximation
because it does not freeze the expert scores at a single reference point.
Instead, the local score at the current iterate is
\begin{equation}
    S_e(\mathbf z)
    =
    -
    \frac{\partial R(\mathbf z)}{\partial z_e},
\end{equation}
which changes as $\mathbf z$ changes.

For the squared reconstruction loss,
Eq.~\eqref{eq:marginal_reconstruction_score} makes this dependence explicit:
both the normalization factor $D_t(\mathbf z)$ and the reconstructed mixture
$\widetilde y_t(\mathbf z)$ depend on the current values of all
expert-selection variables. As optimization changes $\mathbf z$, the
reconstruction residual changes, and the marginal value of each expert is
therefore recomputed relative to the updated mixture.

This captures interactions that are absent from a frozen-score rule. For
example, two experts may both receive large local scores at an initial
reference point because each helps reduce the same component of the
reconstruction residual. After one of them receives a larger selection
weight, that component of the residual may already be substantially
reconstructed, reducing the marginal value of the other expert. Conversely,
an expert with a smaller initial score may become relatively more valuable if
it provides a complementary direction that addresses the residual left by
the currently emphasized experts.

Thus, the distinction between fixed-score selection and OptEP can be
viewed as the distinction between a single frozen local ranking and a
selection-dependent ranking that evolves throughout nonlinear optimization.

\paragraph{Connection to existing score-based pruning}
Methods such as Frequency, EASY-EP, and REAP compute a scalar importance
score for each expert and retain the highest-scoring experts. Although they
differ in how the expert scores are constructed, their final selection rule
has the additive form
\begin{equation}
    \max_{\mathbf z}
    \sum_{e=1}^{E}s_e z_e
    \qquad
    \text{subject to}
    \qquad
    \mathbf z\in\{0,1\}^{E},
    \quad
    \sum_{e=1}^{E}z_e=s.
\end{equation}
The first-order approximation of OptEP has the same fixed-score selection
structure, but with scores derived from the local reconstruction gradient.
The full nonlinear continuous reconstruction optimization goes beyond this
static approximation by allowing the effective expert scores to change as
the reconstructed mixture evolves.

\section{Experiments Details}
\label{app:implementation_details}

\subsection{Calibration Data Details} \label{sec:calibration:details}

\paragraph{Qwen/GLM calibration data} For the Qwen and GLM models, we construct a separate calibration set for each of the three seeds by randomly sampling prompt–completion pairs until reaching the fixed budget of $2^{19}=524{,}288$ tokens. Calibration samples for coding are drawn from Codeforces problems and calibration samples for math are drawn from OpenR1-Math-220k problems, together with their corresponding model-generated completions, including reasoning traces. We limit the calibration samples to correct only responses. We generate these completions using temperature 0.6, top-$p$ 0.95, and a maximum of 32,768 new tokens in both domains. 

Depending on the seed, the fixed token budget corresponds to 48 coding samples for Qwen3-30B-A3B, 43--46 coding samples for Qwen3-235B-A22B,
and 75--93 coding samples for GLM-4.5-Air; for math, it corresponds to 77 samples for Qwen3-30B-A3B, 63 samples for Qwen3-235B-A22B, 
and 49--58 samples for GLM-4.5-Air.

\paragraph{Mixtral-8x7B calibration data} Mixtral-8x7B performs poorly on MATH-500 and LiveCodeBench, so we use easier math and coding benchmarks for both calibration and evaluation. Unlike the Qwen setting, where calibration and evaluation use different datasets, Mixtral-8x7B uses different splits of the same dataset: GSM8K for math and MBPP for coding.

For math, the Mixtral-8x7B calibration pool consists of all 7,473 GSM8K training examples together with completions generated by the full model. We generate completions using temperature 0.0, top-$p$ 1.0, and a maximum of 512 new tokens for both math and coding. For each of five seeds, we randomly sample prompt–completion pairs with correct full-model responses until reaching a fixed budget of 262,144 tokens. This results in 1,248–1,276 GSM8K training examples per calibration set, depending on the seed. 

For coding, we use the full MBPP dataset of 974 examples to construct seed-specific calibration and evaluation splits. For each seed, we generate full-model completions using the same decoding settings and randomly sample prompt--completion pairs until reaching the 262,144-token calibration budget. We then retain only traces for which the full model produces the correct answer, resulting in 267--287 calibration sequences per seed. Examples sampled toward the token budget but yielding incorrect responses are discarded and used in neither calibration nor evaluation. Excluding these examples from evaluation avoids artificially increasing the proportion of incorrect full-model responses in the evaluation set. The remaining MBPP examples form the corresponding evaluation split (details below).

\subsection{Evaluation Details} \label{sec:evaluation:details}

\paragraph{Qwen/GLM evaluation} We evaluate the Qwen and GLM pruned models zero-shot on 500 problems in MATH-500 and 268 problems in LiveCodeBench (LCB v4--v5), using a single generation per problem with temperature $0.6$, top-$p=0.95$, and a maximum generation length of 32,768 tokens. We report mean pass@1 accuracy and standard error across three seeds, where each seed determines both the sampled calibration set and the stochastic model generations during evaluation. Thus, the reported standard error reflects variability from both calibration-set selection and inference-time sampling.

\paragraph{Mixtral-8x7B evaluation} For Mixtral-8x7B math, we evaluate mathematical reasoning on the full GSM8K test set of 1,319 questions. We use a single greedy generation per problem (temperature $0$) with a maximum generation length of 512 tokens and report mean pass@1 accuracy and standard error across five calibration seeds. Because decoding is deterministic, there is no additional inference-time sampling variability. For GSM8K, the standard error therefore reflects only variation in the sampled calibration set. 

For Mixtral-8x7B coding, we evaluate code generation on a held-out MBPP set that is disjoint from the calibration set. MBPP is therefore the only setting in which the evaluation set varies across seeds as well as calibration, since the dataset is not large enough to reserve a separate fixed evaluation set while maintaining the calibration budget. MBPP is evaluated on 477–485 problems per seed rather than a fixed test split.
For each seed, the calibration subset and its complementary evaluation subset are fixed and shared across all compared methods. For MBPP evaluation the reported standard error reflects variation in both the calibration and evaluation subsets, since the calibration seed determines the corresponding held-out evaluation set.

\subsection{Baselines} \label{sec:baselines:details}

We provide additional details on the baselines used:

\begin{itemize}[leftmargin=*, noitemsep]
\item \textbf{Frequency pruning} \citep{muzio2024seer} counts the number of tokens for which each expert is activated and retains the experts with the highest activation counts in each layer.

\item \textbf{Random pruning} randomly selects the subset of experts to retain independently in each layer.

\item \textbf{EASY-EP} \citep{dong2025domain} assigns each expert an importance score based on its router probability, expert output norm, and token-importance weight, where token importance is computed from the cosine similarity between the MoE-layer input and output. Experts with the highest scores are retained.

\item \textbf{REAP} \citep{lasby2025reap} scores each expert using its router weight and activation magnitude, averaged over the calibration tokens for which that expert is active. This reduces the influence of activation frequency and helps retain infrequently used experts that make large contributions when selected. Experts with the highest scores are retained.

\item \textbf{DiEP} \citep{bai2025diep} uses a differentiable, non-uniform pruning strategy in which each expert and each layer are assigned learned importance scores. The expert- and layer-level scores are jointly trained and combined to determine which experts to prune, allowing different numbers of experts to be retained across layers. 

\item \textbf{M-SMoE} \citep{li2023merge} combines expert pruning with expert merging. Rather than simply dropping the pruned experts, it merges experts based on the routing structure, producing a compressed set of experts with updated weights.

\item \textbf{NAEE} \citep{lu2024not} performs layerwise reconstruction-based expert pruning. For each layer, it selects the subset of experts that minimizes the reconstruction error between the original and pruned MoE-layer outputs. The optimization is solved by enumerating all candidate expert subsets, making it practical primarily for MoEs with relatively few experts per layer. Because the Qwen models contain 128 experts per layer, exhaustive enumeration is computationally infeasible. We therefore omit this method for Qwen and evaluate it only on Mixtral-8x7B, which has 8 experts per layer.

\end{itemize}

\subsection{Optimization Details} \label{sec:optimization:details}

\paragraph{Pruning-only optimization}
OptEP (uw), OptEP (w) and OptEP (w-alt) all optimize expert selection \emph{independently} for each layer. For every layer, the optimization uses activations from a forward pass of the full, unpruned model; pruning decisions from earlier layers are therefore not propagated to subsequent layers. We use this independent layerwise setup to ensure a fair comparison with the pruning baselines, which likewise select experts independently across layers.

For the expert-selection step in Eq.~\eqref{eq:continuous_pruning}, we optimize each layer for 50 steps using SGD without momentum and a learning rate of $0.02$. We initialize $z_e=0.5$ for all experts, clip the gradient norm to $5$, and clamp the selection variables to $[0,1]$ after each update. The pretrained router and expert weights remain fixed throughout this stage. We use $\alpha=1$ for Mixtral-8x7B and $\alpha=2$ for both Qwen and GLM models.

\paragraph{Weight update optimization}

For the reconstruction-based weight update in Eq.~\eqref{eq:expert_weight_update}, we optimize only the retained expert weights while keeping the router and all other parameters of the pruned model fixed. Unlike the pruning-only stage, weight updates are performed \emph{sequentially} across layers: after updating a layer, we run the partially updated pruned model forward to generate the inputs used to optimize the next layer.

For Qwen and GLM, we use Adam with a learning rate of $10^{-4}$, gradient-norm clipping at $1$, and $\lambda=0.01$, using the same calibration data as in the pruning stage without a separate validation set. Because the Qwen and GLM calibration datasets contain relatively few samples, we do not hold out a separate validation set to select the number of optimization steps. Instead, we fix the number of steps to either 25 or 50, depending on the setting.
We use 50 optimization steps per layer for the smaller Qwen3-30B-A3B model and 25 steps per layer for the larger Qwen3-235B-A22B and GLM-4.5-Air models.

For Mixtral-8x7B, we optimize the retained expert weights for up to 150 steps using Adam with a learning rate of $3\times10^{-4}$, gradient-norm clipping at $1$, and $\lambda=0.01$. We split the calibration data into 80\% training and 20\% validation subsets, evaluate validation loss every 10 optimization steps, and select the checkpoint with the lowest validation loss for each layer.

\section{Additional Results} \label{sec:additional:results}

We provide the full results for all evaluated models in Tables \ref{tab:qwen30b-math-results}, \ref{tab:qwen30b-code-results}, \ref{tab:qwen235b-math-results}, \ref{tab:qwen235b-code-results}, \ref{tab:glm45air-math-results}, \ref{tab:glm45air-code-results}, and \ref{tab:mixtral_results_full}. These tables report mean performance together with standard errors across experimental seeds and include results for the additional token-weighting mechanisms considered in our experiments.

\subsection{Alternative Weighted Reconstruction Objective: OptEP (w-alt)}

In addition to OptEP (w), which aggregates the inverse activation frequencies of all experts active for a token, we consider an alternative weighting scheme that places greater emphasis on whether a token is routed through a particularly infrequently activated expert. For a token with layer input $\boldsymbol{x}$, we define
\begin{equation}
\label{eq:weights:min}
    w(\boldsymbol{x})
    =
    \frac{1}{
    \min_{e\in\mathcal{T}_K(\boldsymbol{a}(\boldsymbol{x}))} n_e
    }.
\end{equation}
Thus, the weight is determined by the least frequently activated expert among those active for the token. Whereas OptEP (w) accounts for the inverse frequencies of all active experts, OptEP (w-alt) focuses on the rarest active expert, thereby more directly emphasizing tokens involving infrequently activated experts. The full results in Appendix \ref{sec:additional:results} also include this weighting mechanism.

\begin{table}[H]
\centering
\caption{Expert pruning on Qwen3-30B-A3B, evaluated on MATH-500.
Results are mean accuracy $\pm$ standard error over three calibration seeds.}
\label{tab:qwen30b-math-results}
\begin{tabular}{lccc}
\toprule
\multirow{2}{*}{Method}
& \multicolumn{3}{c}{Experts retained per layer} \\
\cmidrule(lr){2-4}
& $s{=}32$ & $s{=}48$ & $s{=}64$ \\
\midrule
Random
& $0.00 \pm 0.00$
& $0.00 \pm 0.00$
& $0.00 \pm 0.00$ \\

Frequency
& $1.40 \pm 0.12$
& $19.27 \pm 4.48$
& $66.33 \pm 7.80$ \\

EASY-EP
& $31.73 \pm 4.68$
& \underline{$93.73 \pm 0.13$}
& $94.60 \pm 0.40$ \\

REAP
& $76.07 \pm 3.21$
& $91.93 \pm 0.58$
& $94.73 \pm 0.24$ \\

M-SMoE
& $1.33 \pm 0.29$
& $1.00 \pm 0.31$
& $1.07 \pm 0.41$ \\
\midrule

OptEP (uw)
& \underline{$88.40 \pm 0.50$}
& $93.40 \pm 0.20$
& $94.67 \pm 0.29$ \\

OptEP (w)
& $83.47 \pm 1.44$
& $93.53 \pm 0.55$
& $95.47 \pm 0.35$ \\

OptEP (w-alt)
& $84.47 \pm 2.19$
& $93.40 \pm 0.31$
& \underline{$95.67 \pm 0.53$} \\

OptEP (uw) + EWU
& $\mathbf{90.20 \pm 0.53}$
& $93.27 \pm 0.58$
& $95.33 \pm 0.41$ \\

OptEP (w) + EWU
& $87.40 \pm 2.10$
& $\mathbf{94.40 \pm 0.12}$
& $\mathbf{96.13 \pm 0.07}$ \\

\midrule\midrule

Full model
& \multicolumn{3}{c}{$96.40 \pm 0.12$} \\

\bottomrule
\end{tabular}
\end{table}

\begin{table}[H]
\centering
\caption{Expert pruning on Qwen3-30B-A3B, evaluated on LiveCodeBench v4--v5.
Results are mean accuracy $\pm$ standard error over three calibration seeds.}
\label{tab:qwen30b-code-results}
\begin{tabular}{lccc}
\toprule
\multirow{2}{*}{Method}
& \multicolumn{3}{c}{Experts retained per layer} \\
\cmidrule(lr){2-4}
& $s{=}48$ & $s{=}64$ & $s{=}96$ \\
\midrule
Random
& $0.00 \pm 0.00$
& $0.00 \pm 0.00$
& $0.00 \pm 0.00$ \\

Frequency
& $0.00 \pm 0.00$
& $1.87 \pm 0.43$
& $\mathbf{61.44 \pm 0.25}$ \\

EASY-EP
& $17.54 \pm 1.92$
& $54.73 \pm 0.97$
& $61.07 \pm 0.69$ \\

REAP
& $38.43 \pm 1.68$
& $52.49 \pm 0.76$
& $59.83 \pm 1.06$ \\

M-SMoE
& $0.00 \pm 0.00$
& $0.00 \pm 0.00$
& $59.83 \pm 0.90$ \\
\midrule

OptEP (uw)
& $50.50 \pm 1.63$
& $56.22 \pm 0.82$
& $60.57 \pm 1.19$ \\

OptEP (w)
& $48.88 \pm 0.37$
& $54.60 \pm 1.96$
& $\mathbf{61.44 \pm 1.43}$ \\

OptEP (w-alt)
& $48.13 \pm 1.63$
& $55.85 \pm 1.37$
& $59.08 \pm 0.66$ \\

OptEP (uw) + EWU
& \underline{$52.11 \pm 1.86$}
& $\mathbf{57.71 \pm 0.25}$
& $60.70 \pm 0.33$ \\

OptEP (w) + EWU
& $\mathbf{53.48 \pm 0.69}$
& \underline{$56.59 \pm 1.32$}
& \underline{$61.19 \pm 0.75$} \\
\midrule\midrule

Full model
& \multicolumn{3}{c}{$60.57 \pm 0.45$} \\

\bottomrule
\end{tabular}
\end{table}

\begin{table}[H]
\centering
\caption{Expert pruning on Qwen3-235B-A22B, evaluated on MATH-500.
Results are mean accuracy $\pm$ standard error over three calibration seeds.}
\label{tab:qwen235b-math-results}
\begin{tabular}{lccc}
\toprule
\multirow{2}{*}{Method}
& \multicolumn{3}{c}{Experts retained per layer} \\
\cmidrule(lr){2-4}
& $s{=}32$ & $s{=}48$ & $s{=}64$ \\
\midrule
Random
& $0.67 \pm 0.29$
& $1.13 \pm 0.07$
& $0.60 \pm 0.00$ \\

Frequency
& $0.07 \pm 0.07$
& $0.27 \pm 0.07$
& $93.20 \pm 0.46$ \\

EASY-EP
& $71.33 \pm 7.64$
& $93.20 \pm 0.92$
& $95.20 \pm 0.50$ \\

REAP
& $84.07 \pm 2.42$
& $93.60 \pm 0.23$
& $96.07 \pm 0.37$ \\

M-SMoE
& $0.67 \pm 0.29$
& $0.13 \pm 0.07$
& $60.07 \pm 13.75$ \\
\midrule

OptEP (uw)
& $87.80 \pm 0.53$
& $92.53 \pm 1.38$
& $95.53 \pm 0.29$ \\

OptEP (w)
& \underline{$91.87 \pm 0.58$}
& $95.27 \pm 0.24$
& $96.13 \pm 0.29$ \\

OptEP (w-alt)
& $91.53 \pm 0.55$
& $94.53 \pm 0.66$
& \underline{$96.20 \pm 0.53$} \\

OptEP (uw) + EWU
& $90.53 \pm 1.12$
& \underline{$95.53 \pm 0.35$}
& \underline{$96.20 \pm 0.20$} \\

OptEP (w) + EWU
& $\mathbf{94.67 \pm 0.13}$
& $\mathbf{96.13 \pm 0.29}$
& $\mathbf{96.47 \pm 0.35}$ \\
\midrule\midrule

Full model
& \multicolumn{3}{c}{$97.60 \pm 0.00$} \\

\bottomrule
\end{tabular}
\end{table}

\begin{table}[H]
\centering
\caption{Expert pruning on Qwen3-235B-A22B, evaluated on LiveCodeBench v4--v5.
Results are mean accuracy $\pm$ standard error over three calibration seeds.}
\label{tab:qwen235b-code-results}
\begin{tabular}{lccc}
\toprule
\multirow{2}{*}{Method}
& \multicolumn{3}{c}{Experts retained per layer} \\
\cmidrule(lr){2-4}
& $s{=}48$ & $s{=}64$ & $s{=}96$ \\
\midrule
Random
& $0.00 \pm 0.00$
& $0.00 \pm 0.00$
& $0.00 \pm 0.00$ \\

Frequency
& $0.00 \pm 0.00$
& $45.52 \pm 2.06$
& $64.43 \pm 2.37$ \\

EASY-EP
& $41.67 \pm 4.21$
& $56.97 \pm 0.33$
& $61.19 \pm 1.41$ \\

REAP
& $41.17 \pm 2.61$
& $61.69 \pm 1.73$
& $64.68 \pm 1.40$ \\

M-SMoE
& $0.00 \pm 0.00$
& $27.11 \pm 2.52$
& $64.68 \pm 1.76$ \\
\midrule

OptEP (uw)
& $43.04 \pm 2.78$
& $62.31 \pm 0.78$
& $65.17 \pm 0.66$ \\

OptEP (w)
& $51.99 \pm 1.08$
& $62.44 \pm 0.82$
& $64.18 \pm 1.41$ \\

OptEP (w-alt)
& $51.37 \pm 2.16$
& $61.94 \pm 0.65$
& $\mathbf{66.67 \pm 0.90}$ \\

OptEP (uw) + EWU 
& \underline{$54.85 \pm 1.12$}
& $\mathbf{63.31 \pm 0.70}$
& $65.30 \pm 1.14$ \\

OptEP (w) + EWU
& $\mathbf{56.22 \pm 1.43}$
& \underline{$62.94 \pm 2.61$}
& \underline{$66.42 \pm 1.51$} \\
\midrule\midrule

Full model
& \multicolumn{3}{c}{$66.67 \pm 0.66$} \\

\bottomrule
\end{tabular}
\end{table}

\begin{table}[H]
\centering
\caption{Expert pruning on GLM-4.5-Air, evaluated on MATH-500.
Results are mean accuracy $\pm$ standard error over three calibration seeds.}
\label{tab:glm45air-math-results}
\begin{tabular}{lccc}
\toprule
\multirow{2}{*}{Method}
& \multicolumn{3}{c}{Experts retained per layer} \\
\cmidrule(lr){2-4}
& $s{=}32$ & $s{=}48$ & $s{=}64$ \\
\midrule
Random
& $1.40 \pm 0.12$
& $5.13 \pm 0.64$
& $27.07 \pm 6.89$ \\

Frequency
& $26.47 \pm 1.16$
& $56.93 \pm 4.91$
& $74.87 \pm 3.96$ \\

EASY-EP
& $71.80 \pm 2.66$
& $92.07 \pm 1.07$
& $95.53 \pm 0.79$ \\

REAP
& $75.87 \pm 2.03$
& $93.73 \pm 0.07$
& $96.60 \pm 0.31$ \\

M-SMoE
& $2.47 \pm 0.58$
& $27.53 \pm 2.49$
& $67.27 \pm 3.46$ \\
\midrule

OptEP (uw)
& $77.73 \pm 1.84$
& $92.53 \pm 1.09$
& $\mathbf{97.07 \pm 0.07}$ \\

OptEP (w)
& $80.47 \pm 0.64$
& $93.73 \pm 0.07$
& $96.20 \pm 0.40$ \\

OptEP (w-alt)
& $77.27 \pm 0.66$
& $93.33 \pm 0.18$
& $96.47 \pm 0.18$ \\

OptEP (uw) + EWU
& \underline{$85.93 \pm 0.85$}
& \underline{$94.20 \pm 0.12$}
& \underline{$97.00 \pm 0.20$} \\

OptEP (w) + EWU
& $\mathbf{86.93 \pm 0.98}$
& $\mathbf{94.93 \pm 0.24}$
& \underline{$97.00 \pm 0.12$} \\

\midrule\midrule

Full model
& \multicolumn{3}{c}{$97.87 \pm 0.58$} \\

\bottomrule
\end{tabular}
\end{table}

\begin{table}[H]
\centering
\caption{Expert pruning on GLM-4.5-Air, evaluated on LiveCodeBench v4--v5.
Results are mean accuracy $\pm$ standard error over three calibration seeds.}
\label{tab:glm45air-code-results}
\begin{tabular}{lccc}
\toprule
\multirow{2}{*}{Method}
& \multicolumn{3}{c}{Experts retained per layer} \\
\cmidrule(lr){2-4}
& $s{=}64$ & $s{=}80$ & $s{=}96$ \\
\midrule
Random
& $0.50 \pm 0.25$
& $3.48 \pm 0.69$
& $22.39 \pm 4.93$ \\

Frequency
& $3.24 \pm 0.25$
& $14.68 \pm 1.51$
& $32.21 \pm 2.00$ \\

EASY-EP
& $45.77 \pm 1.67$
& $54.85 \pm 0.78$
& \underline{$64.30 \pm 1.08$} \\

REAP
& $49.88 \pm 0.54$
& $59.33 \pm 1.41$
& $64.18 \pm 0.37$ \\

M-SMoE
& $1.24 \pm 0.12$
& $9.33 \pm 0.86$
& $27.49 \pm 1.59$ \\
\midrule

OptEP (uw)
& $51.99 \pm 1.39$
& $62.56 \pm 2.00$
& $62.06 \pm 2.36$ \\

OptEP (w)
& $52.61 \pm 0.78$
& $62.69 \pm 1.55$
& \underline{$64.30 \pm 0.69$} \\

OptEP (w-alt)
& $53.73 \pm 0.43$
& \underline{$63.06 \pm 0.99$}
& \underline{$64.30 \pm 0.12$} \\

OptEP (uw) + EWU
& \underline{$54.85 \pm 0.37$}
& $\mathbf{63.56 \pm 0.82}$
& $\mathbf{64.31 \pm 0.33}$ \\

OptEP (w) + EWU
& $\mathbf{56.22 \pm 1.74}$
& $61.82 \pm 1.00$
& $62.06 \pm 1.95$ \\

\midrule\midrule

Full model
& \multicolumn{3}{c}{$65.30 \pm 0.21$} \\

\bottomrule
\end{tabular}
\end{table}

\begin{table}[H]
\centering
\caption{Expert pruning results for Mixtral-8x7B-v0.1 on math (GSM8K flexible extraction) and coding (MBPP standard pass@1). Results are mean accuracy $\pm$ standard error in \% over five seeds.}
\label{tab:mixtral_results_full}
\begin{tabular}{lcccc}
\toprule
\multirow{2}{*}{Method}
& \multicolumn{2}{c}{Math (GSM8K)}
& \multicolumn{2}{c}{Coding (MBPP)} \\
\cmidrule(lr){2-3}\cmidrule(lr){4-5}
& $s{=}4$ & $s{=}6$ & $s{=}4$ & $s{=}6$ \\
\midrule
Random
& $3.08 \pm 1.44$
& $20.15 \pm 2.13$
& $7.74 \pm 4.68$
& $40.25 \pm 0.66$ \\

Frequency
& $8.75 \pm 0.48$
& $8.96 \pm 0.24$
& $18.71 \pm 0.40$
& $37.51 \pm 0.60$ \\

EASY-EP
& $7.01 \pm 0.24$
& $25.26 \pm 0.15$
& $28.70 \pm 0.96$
& $47.57 \pm 0.34$ \\

REAP
& $11.45 \pm 0.79$
& $23.55 \pm 0.55$
& $25.07 \pm 0.72$
& $44.49 \pm 0.21$ \\

M-SMoE
& $10.63 \pm 2.33$
& $30.51 \pm 0.38$
& $24.45 \pm 0.65$
& $38.05 \pm 0.38$ \\

NAEE brute-force
& $11.96 \pm 0.47$
& $24.96 \pm 0.06$
& $28.53 \pm 0.47$
& $43.87 \pm 0.47$ \\

DiEP
& $10.19 \pm 0.27$
& $24.90 \pm 0.69$
& $16.54 \pm 0.91$
& $41.91 \pm 0.60$ \\
\midrule

OptEP (uw)
& $14.04 \pm 0.74$
& $24.58 \pm 0.45$
& $27.99 \pm 0.63$
& $47.48 \pm 0.32$ \\

OptEP (w)
& $12.89 \pm 0.00$
& $25.32 \pm 0.37$
& $30.31 \pm 0.61$
& $47.61 \pm 0.59$ \\

OptEP (w-alt)
& $14.54 \pm 0.80$
& $22.75 \pm 0.33$
& $28.02 \pm 0.31$
& \underline{$48.07 \pm 0.19$} \\

OptEP (uw) + EWU
& $\mathbf{30.42 \pm 0.56}$
& $\mathbf{33.48 \pm 0.54}$
& \underline{$39.37 \pm 0.77$}
& $47.61 \pm 0.72$ \\

OptEP (w) + EWU
& \underline{$28.16 \pm 0.54$}
& \underline{$32.90 \pm 0.57$}
& $\mathbf{39.50 \pm 0.33}$
& $\mathbf{50.19 \pm 0.56}$ \\
\midrule

Brute Force: OptEP (uw) obj
& $12.37 \pm 0.32$
& $21.62 \pm 0.44$
& $28.56 \pm 0.21$
& $46.40 \pm 0.39$ \\

Brute Force: OptEP (w) obj
& $14.63 \pm 1.05$
& $23.05 \pm 0.54$
& $27.78 \pm 0.45$
& $45.86 \pm 0.45$ \\

\midrule\midrule

Full model (8 experts)
& \multicolumn{2}{c}{$47.61 \pm 0.00$}
& \multicolumn{2}{c}{$59.34 \pm 0.86$} \\

\bottomrule
\end{tabular}
\end{table}

\end{document}